\documentclass[conference]{IEEEtran}
\IEEEoverridecommandlockouts
\usepackage{cite}
\usepackage{amsmath,amssymb,amsfonts}
\usepackage{algorithmic}
\usepackage{graphicx}
\usepackage{textcomp}
\usepackage{xcolor}
\usepackage{lipsum}
\usepackage{array}
\usepackage{booktabs}
\usepackage{tabularx}
\usepackage{multicol}
\usepackage{multirow}
\usepackage{xcolor}
\usepackage{amssymb}
\usepackage{amsmath}
\usepackage{stfloats}
\usepackage{mathdots}
\usepackage{graphicx} 
\usepackage{caption}
\usepackage{subfigure}
\usepackage{algorithm}
\def\BibTeX{{\rm B\kern-.05em{\sc i\kern-.025em b}\kern-.08em
    T\kern-.1667em\lower.7ex\hbox{E}\kern-.125emX}}

\begin{document}

\title{Fast-Spanning Ant Colony Optimisation (FaSACO) for Mobile Robot Coverage Path Planning\\
\thanks{This paper was supported by the Manchester Metropolitan University CfACS seed project: Machine Learning for Affordable Mobile Robots.}
}

\author{\IEEEauthorblockN{1\textsuperscript{st} Christopher Carr}
\IEEEauthorblockA{\textit{Department of Computing and Mathematics} \\
\textit{Manchester Metropolitan University}\\
Manchester, UK \\
christopher.carr@stu.mmu.ac.uk}
\and
\IEEEauthorblockN{2\textsuperscript{nd} Peng Wang}
\IEEEauthorblockA{\textit{Department of Computing and Mathematics} \\
\textit{Manchester Metropolitan University}\\
Manchester, UK  \\
p.wang@mmu.ac.uk}
}

\maketitle

\begin{abstract}
Coverage Path Planning (CPP) aims at finding an optimal path that covers the whole given space. Due to the NP-hard nature, CPP remains a challenging problem. Bio-inspired algorithms such as Ant Colony Optimisation (ACO) have been exploited to solve the problem because they can utilise heuristic information to mitigate the path planning complexity. This paper proposes the Fast-Spanning Ant Colony Optimisation (FaSACO), where ants can explore the environment with various velocities. By doing so, ants with higher velocities can find destinations or obstacles faster and keep lower velocity ants informed by communicating such information via pheromone trails on the path. This mechanism ensures that the (sub-)~optimal path is found while reducing the overall path planning time. Experimental results show that FaSACO is $19.3-32.3\%$ more efficient than ACO in terms of CPU time, and re-covers $6.9-12.5\%$ less cells than ACO. This makes FaSACO appealing in real-time and energy-limited applications.

\end{abstract}


\section{Introduction}
Coverage path planning aims at finding a path that can completely cover a given space, which is required by a lot of robot-based applications~\cite{nasirian2021efficient,almadhoun2016survey}. For instance, the adoption of an autonomous robot disinfector helps lower transmission risks to patients and medical personnel \cite{nasirian2021efficient}. The inspection of indoor and outdoor structures by aerial robots avoids exposing human operators to hazardous scenarios \cite{almadhoun2016survey}. Similar applications include robot painters, damming robots, lawn mowers, automated harvesters, and window cleaners \cite{galceran2013survey}.

Despite the variety of applications, CPP remains a challenge due to the fact that it is an NP-hard problem. Essentially, CPP resembles the covering salesman problem (CSP), which is a variant of the traveling salesman problem (TSP). In robotics, TSP requires a robot to visit each node of a graph that represents an environment once and with the minimum total cost. It is well-known that the TSP/CSP problem becomes computationally inefficient when the number of cities increases~\cite{arkin1994approximation}. Similar to the CSP problem, the complexity of CPP is determined by the size of the environment. For a large-scale environment, its map is usually decomposed into smaller sub-areas, and CPP is then performed in each sub-area. Boustrophedon decomposition~\cite{choset2001coverage} is one of the most popular decomposition methods, where free space is divided into smaller sub-areas by sweeping a line through the whole map in one direction\cite{nasirian2021efficient}. All the sub-areas are next covered in a TSP manner to gradually cover the whole map completely.

\begin{figure}[t]
    \centering
    \includegraphics[width=1.0\linewidth]{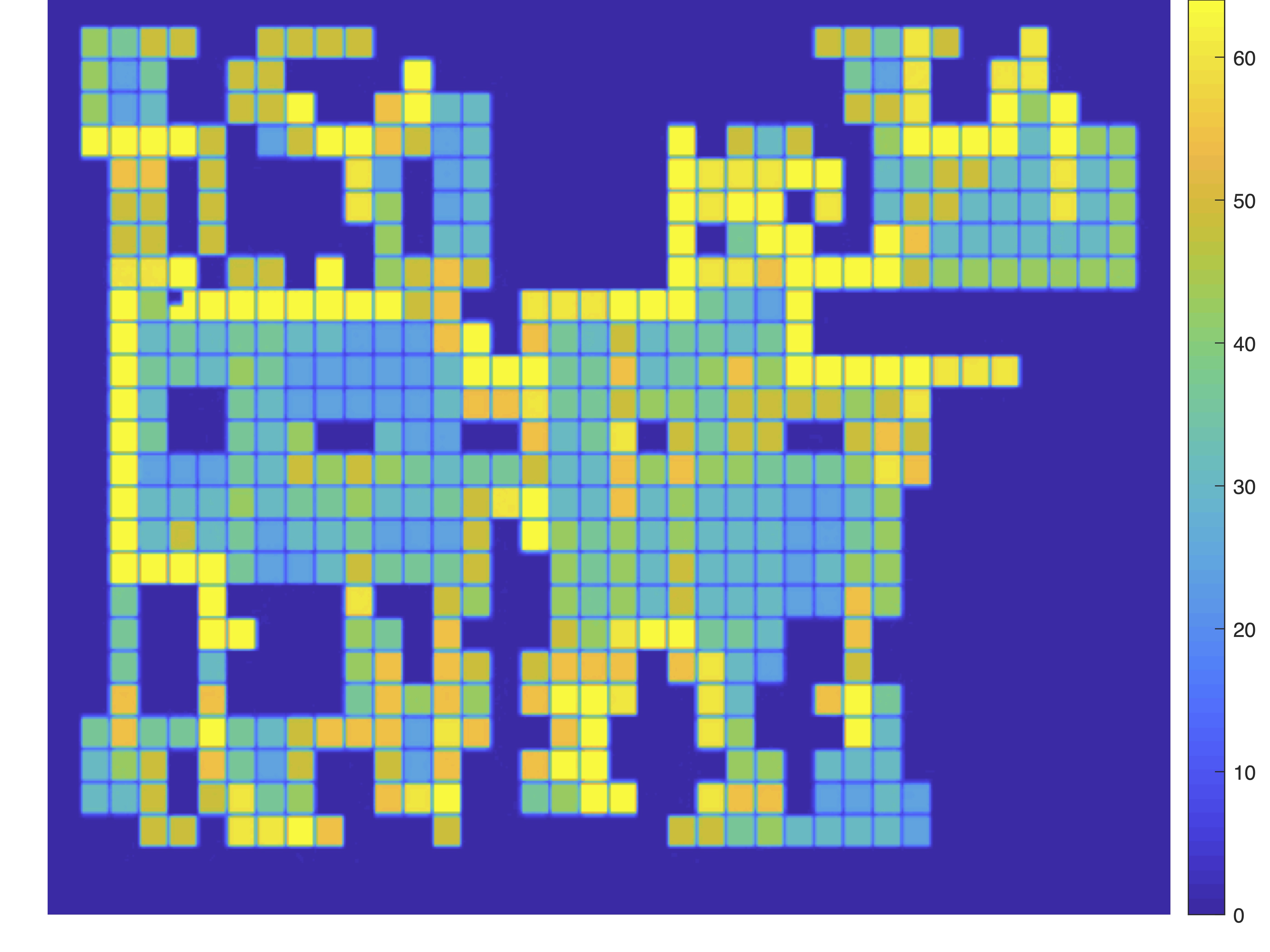}
    \caption{Pheromone Trail Generated by Ants from FaSACO at a Time Instant on an Occupancy Grip Map. The pheromone trail is generated by ants who have conducted CPP up to now, and it will serve as heuristic information for ants who haven't conducted CPP. This pheromone trail helps ants to find the optimal coverage path more efficiently.}
    \label{fig:phromone}
\end{figure}

The literature has seen a rich amount of CPP algorithms, and one popular taxonomy is to divide them into classical algorithms and heuristic-based algorithms \cite{tan2021comprehensive} depending on whether heuristic information is used. Tan et al. have provided a comprehensive survey in~\cite{tan2021comprehensive}, where popular classical algorithms such as the Spiral Spanning Tree Coverage (Spiral-STC), ZigZag and heuristic algorithms such as Ant Colony Optimisation (ACO) are discussed. Spiral-STC was proposed by Gabriely and Rimon~\cite{gabriely2002spiral}, where the free space of an environment was decomposed into cells that were structured by a spanning tree. By performing tree search algorithms such as depth-first search, a path would be found for the robot to cover the whole free space. In comparison, ZigZag algorithm starts with decomposing the free space into cells as well~\cite{gao2020frontier}, and then follows a zigzag, `mowing the lawn' pattern to cover all the free cells~\cite{galceran2013survey}.  In general, traditional coverage algorithms lack the `intelligence' to exploit heuristic information to help solve CPP problems~\cite{bormann2018indoor}. In contrast, ACO can utilise heuristic information such as pheromone trails to optimise path planning, but the swarm intelligence nature makes it time costly. In the current research of ACO for CPP, a constant velocity is assigned to all ants, which limits the diversity of the ant colony~\cite{tan2021comprehensive}. Nevertheless, the potential to generate optimal path and acceleration with parallel computation still makes ACO appealing.

In this paper, we propose the Fast-Spanning ACO (FaSACO) where ants are of various velocities compared to ACO, mimicking the reality that some ants move faster than others. In such a manner, faster-moving ants are more likely to find destinations or obstacles, and will communicate such information to slower ants via pheromone trails on the path, and slower ants can then optimise their path based on such pheromone trails. Figure \ref{fig:phromone} shows the pheromone trails generated by FaSACO at a certain time instant. One can see the variation of pheromones indicating the preferences of cells by ants (the pheromones are contributed by ants conducted CPP up to now, and the higher pheromone on a cell the more likely ants will visit it). In brief, the contributions of this paper include 1) Enabling ants in FaSACO to move with different velocities; 2) FaSACO outperforms ACO both in terms of the number of re-covered cells (cells that are visited twice or more times by the robot) and CPU time; 3) A series of experiments were conducted to compare FaSACO with ACO, Sprial-STC, and ZigZag, to demonstrate that FaSACO can achieve a trade-off between the number of re-covered cells and the CPU time. 

The remaining part of the paper is structured as follows: Section II introduces the  environment and motion models used for CPP in the paper, Section III proposes and elaborates upon the fast-spanning ACO, Section IV details experimental results and analysis, and the paper is concluded in Section V.

\section{Environment Model and Robot Motion Model}
\begin{figure*}[hb]
    \centering
    \includegraphics[width=0.62\linewidth]{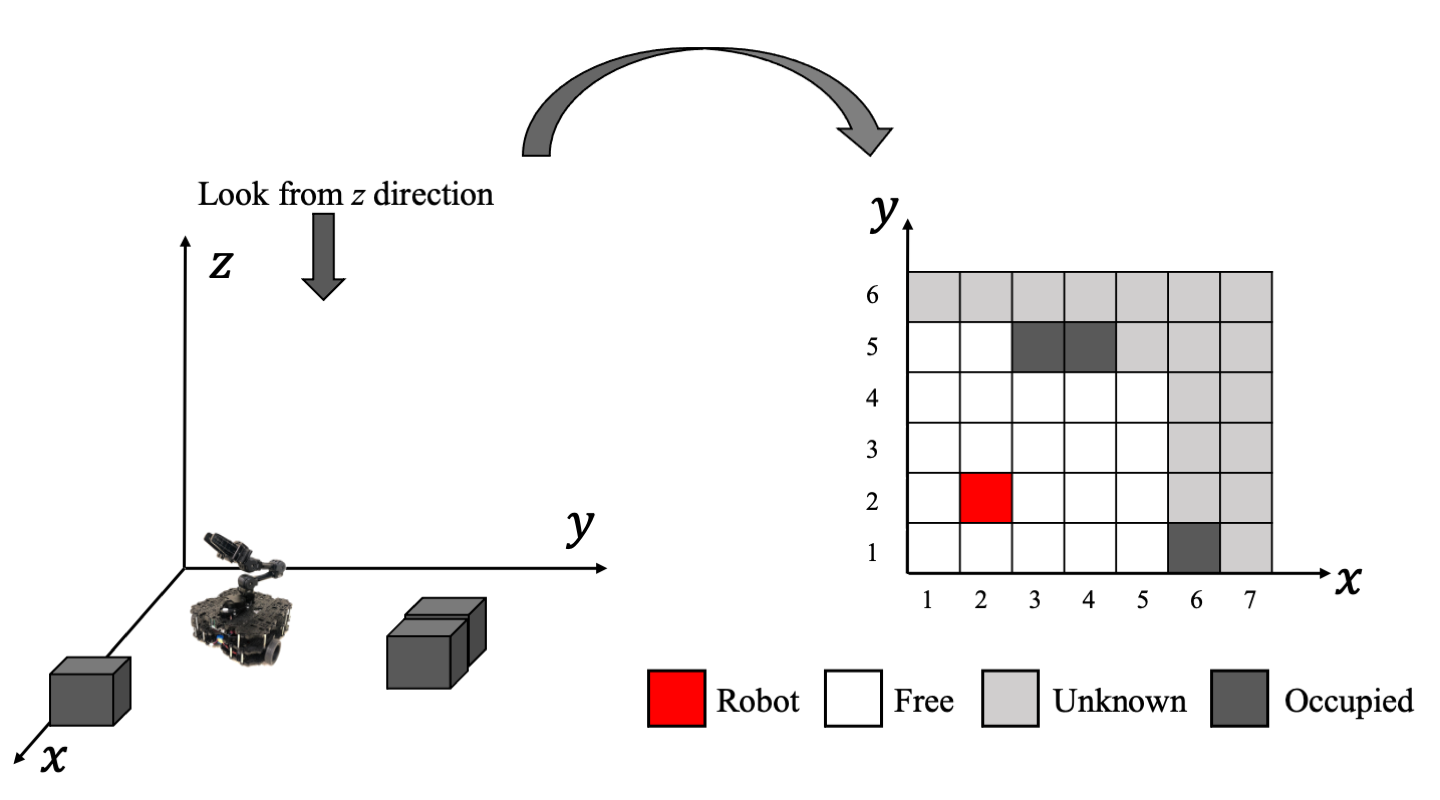}
    \caption{An Example Environment and the Corresponding Occupancy Grid Map}
    \label{fig:occ_grid}
\end{figure*}
\subsection{Environment Model}
The environment model provides a robot with a description of its surroundings. For instance, a robot disinfector or lawn mower can plan a path based on the map of a given environment to cover the whole environment completely and efficiently. This paper uses the occupancy grid map as the environment model, which is detailed as follows.

For an environment $E$, the corresponding occupancy grid map $M(E)$, or simply $M$, is defined as 
\begin{equation}
    M = \{m_{i,j} |p(m_{i,j}) \in [0, 1], 1 \leq i \leq m_r, 1 \leq j \leq m_c\},
\end{equation}
\noindent
where $M$ is an $m_r \times m_c$ dimensional grid map with resolution $\Delta M$, and a cell $m_{i,j}$ is uniquely indexed by its row $i$ and column $j$ in $M$. The probability of cell $m_{i,j}$ being occupied by obstacles is denoted by $p(m_{i,j})$, which determines the state of a cell according to

\begin{equation} \label{eq:cellstate}
p(m_{i,j})=
\left\{
\begin{array}{lr}
0, &\text{Free or} \: M_f\\
1, &\text{Occupied or} \: M_o\\
0.5, &\text{Unknown or} \: M_u\\
p\in (0,1)\backslash 0.5, &\text{Occupied with Probability} \: p
\end{array}
\right.
\end{equation}
\noindent
For simplicity, a cell $m_{i,j}$ can be denoted as $m_u$, with index 
\begin{equation}
  u=(j-1)*m_c + m_r, 1\leq u \leq m_r*m_c.  
\end{equation}

Figure \ref{fig:occ_grid} shows an example environment with a robot and obstacles in it (left), and a possible occupancy grid map (right) built so far by the robot. We can see that there are cells marked as occupied, e.g., cell $m_{31}$ ($m_{3,5}$) is free, cell $m_{23}$ ($m_{2,4}$) is occupied, and cell $m_{36}$ ($m_{1,6}$) is unknown. By performing CPP, the robot will plan a path to cover all the free cells on the occupancy grid map.

\subsection{Motion Model for Path Planning}

The motion model of a real robot depends on a few factors from the structure of the robot to the motors used. For instance, a differential motion model can be found in~\cite{wang2021feature}. With such a model, a robot can move in an infinite number of directions. 

\begin{figure}[t]
    \centering
    \includegraphics[width=0.6\linewidth]{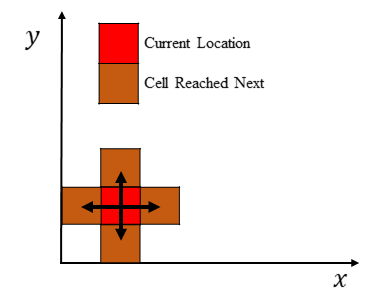}
    \caption{The Robot Motion Model}
    \label{fig:motion}
\end{figure}

In occupancy grid map-based CPP, the motion model is usually simplified in the planning stage such that the robot is restricted to moving in a finite number of directions. Figure \ref{fig:motion} shows such a simplified motion model, where the robot is restricted to moving in four directions relative to its current location: up, down, left, and right. The velocity $v$ in each direction is set to be the same, usually $v$ is set to 1, meaning that the robot moves one cell in each step.

\section{Fast-Spanning Ant Colony Optimisation}
\subsection{Ant Colony Optimisation}

Ant colony optimisation is inspired by the swarm intelligence demonstrated by ants while they are seeking food. By leaving pheromone trails on paths, an ant can communicate with other ants about whether a path leads to food. If the path leads to food, then more ants will follow the path, leaving more pheromones and increasing the pheromone intensity on that path. On the contrary, paths that do not lead to food will attract fewer ants until no pheromone is accumulated, and the paths are eventually discarded. This mechanism has inspired the development of a series of algorithms with applications in optimisation~\cite{dorigo1996ant} and mobile robot path planning~\cite{dai2019mobile}. 

The formulation of ACO varies from case to case. In occupancy grid map-based CPP, the ant colony is composed of $K$ ants in total, denoted as $A= \{a_1, a_2, \cdots,a_k,\cdots a_K\}$. Each ant will move between cells to find a path that covers the whole free space.

Assume ant $a_k$ has visited some cells denoted as $M_k$, and it currently perceives that the robot locates at cell $m_u$, and it will reach cell $m_t\notin M_k$ in the next step. Obviously, the choice of $m_t$ is determined by the index $t$, which is further constrained by the motion model like the one shown in Figure \ref{fig:motion}. Based on such information, ant $a_k$ chooses $m_t$ or simply $t$ based on equation (\ref{eq:ACO}). Note that $t$ has four different options, i.e., $u_1$, $u_2$, $u_3$, and $u_4$ encoding cells above and below, and to the left and right of the current cell $m_u$.
\begin{equation}\label{eq:ACO}
t \leftarrow
\left\{
    \begin{array}{ll}
        \arg\max_{m_t\notin M_k}\Big\{\tau(m_u,m_t)[\eta(m_u,m_t)]^\beta \Big\}, &\text{if} \: q\leq q_0 \\
        \max_{m_t\notin M_k} p_k(m_u,m_t),  &\text{otherwise}\\
    \end{array}
\right.    
\end{equation}
\noindent
where $\tau(m_u,m_t)$ is the amount of pheromone on the path from cell $m_u$ to cell $m_t$, and $\eta(m_u,m_t)$ is the heuristic function, which is defined as 
\begin{equation}
    \eta(m_u,m_t) = \frac{1}{dist(m_u,m_t)},
\end{equation}
\noindent where $dist(\cdot,\cdot)$ is the Manhattan distance between two cells. When the motion model in Figure \ref{fig:motion} is used, $\eta(m_u,m_t)$ will always be 1.
$\beta$ is a constant defining the relative importance of pheromone trail and closeness, $q$ is a random number between 0 and 1, $q_0$ is a threshold constant between 0 and 1, and $p_k(m_u,m_t)$ is used to check reachable cells for the next step. It is designed in favour of adjacent cells $m_u$ with the highest level of pheromone trail:
\begin{equation}\label{eq:ACO1}
p_k(m_u,m_t)=
\left\{
    \begin{array}{ll}
        \frac{\tau(m_u,m_t)[\eta(m_u,m_t)]^\beta}{\sum_{m_t\notin M_k}\tau(m_u,m_t)[\eta(m_u,m_t)]^\beta}, &\text{if} \: m_t\notin M_k \\
        0,  &\text{otherwise}.\\
    \end{array}
\right.    
\end{equation}
\noindent
Note that the `$\leftarrow$' in equation (\ref{eq:ACO}) means mapping probability to cell indices, which are $u_1$, $u_2$, $u_3$, and $u_4$ in this case.

After an ant $a_k$ moves from cell $m_u$ to cell $m_t$, the pheromone on the path will be updated following
\begin{equation}\label{eq:localPheromone}
    \tau(m_u,m_t) = (1-\alpha)\cdot\tau(m_u,m_t) + \alpha \cdot \tau_0,
\end{equation}
\noindent
where $\alpha$ is a constant defining the relative importance of the shortest path and the existing pheromone level and $\tau_0$ is a system parameter.

After all ants finish exploring the environment, the ant finds the shortest path will deposit extra pheromone along the path following
\begin{equation}\label{eq:globalPheromone}
    \varphi(m_u,m_t) = (1-\alpha)\cdot\varphi(m_u,m_t) + \alpha \cdot \Delta\varphi(m_u,m_t),
\end{equation}
\noindent
where $\varphi(m_u,m_t)$ denotes the pheromone on the path from cell $m_u$ to cell $m_t$ upon the completion of all ants, and $\Delta\varphi(m_u,m_t)$ is the amount of pheromone deposited on the shortest path between cell $m_u$ and cell $m_t$, and $\Delta\varphi(m_u,m_t)$ is proportional to the inverse of the shortest distance.

Generally, ACO can find the desired path~\cite{dai2019mobile} by running the $K$ ants for a few iterations. In each iteration, equations (\ref{eq:ACO}) to (\ref{eq:localPheromone}) will be used for updating pheromone trails. After each iteration, there will be a `best' ant that generates the shortest path. Extra pheromones will be deposited along this path following equation (\ref{eq:globalPheromone}) to make it more beneficial for later iterations.

Nevertheless, ACO has been known for its inefficiency due to the `swarm' nature. This has caused the lack of investigation into how other factors affect the efficiency of ACO, e.g., ant velocities in ACO are constant, which does not agree with the reality that ants could move with various velocities. This has inspired us to investigate the effect of ant velocities on the efficiency of ACO, and propose FaSACO as below.

\subsection{Fast-Spanning Ant Colony Optimisation}
Compared to ACO, the velocities of ants in FaSACO are no longer constant. For the colony $A= \{a_1, a_2, \cdots,a_k,\cdots a_K\}$, FaSACO splits the colony into $C$ cohorts
\begin{equation}
   A = \{A_1, A_2, \cdots, A_c, \cdots, A_C\}, 
\end{equation}
\noindent
with $A_c = \{a_{l_1},a_{l_2}, \cdots, a_{l_c}, \cdots, a_{L_c}\}$, $L_c$ the number of ants in the cohort, and $\sum_{c=1}^C |A_c|= K$, where $|.|$ denotes the cardinal number of a set. Next, a velocity $c$ satisfies $1\leq c \leq C$ is set to the cohort $A_c$. By doing so, ants with a greater velocity will move faster in the free space, and update pheromone trails of more cells in each step. One direct consequence will be that they find the destinations or hit into obstacles first. Such information will be communicated to slower ants through the pheromone trails, such that slower ants can use the information to optimise their path.

\begin{figure}[t]
    \centering
    \includegraphics[width=1.0\linewidth]{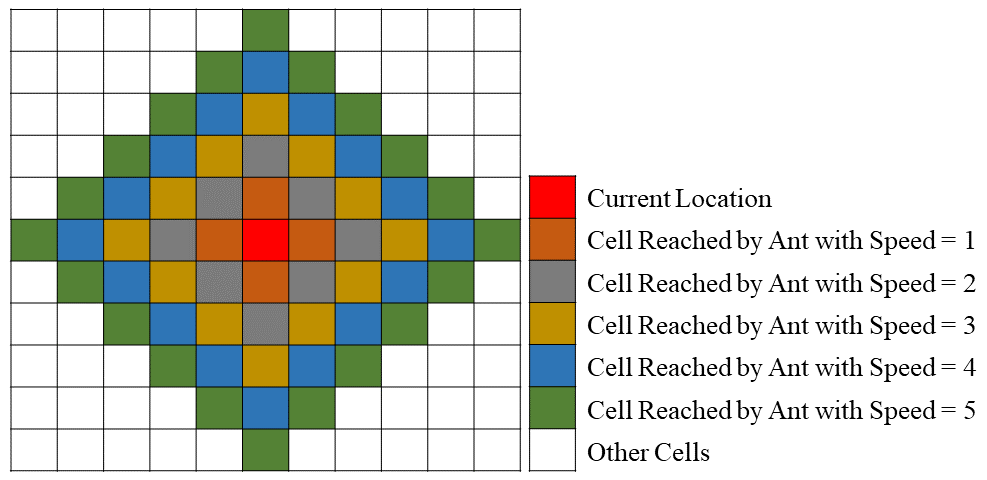}
    \caption{Cells Reached by Ants of Different Velocities}
    \label{fig:ant_speed}
\end{figure}

Figure \ref{fig:ant_speed} shows the fringe cells reached by ants of different velocities. Note that it is obtained based on the motion model shown in Figure \ref{fig:motion}. We can see that ants with greater velocities can explore a wider area compared to ants with smaller velocities.

In FaSACO, equations (\ref{eq:ACO}) to (\ref{eq:globalPheromone}) are still used for generating the next cells to visit, updating local and global pheromones, and getting the best CPP path, etc. However, ants now have different velocities, as detailed in Algorithm \ref{alg:FaSACO}.

\begin{algorithm}[b]
	\caption{FaSACO for Coverage Path Planning}
	\label{alg:FaSACO}
	\begin{algorithmic}[1]
		\renewcommand{\algorithmicrequire}{\textbf{Input:}}
		\REQUIRE A grid map, a motion model, current cell $m_u$, next cell $m_t$, an ant colony divided into cohorts $\{A_1, A_2, \cdots, A_c, \cdots, A_C\}$, with $A_c$ the velocity $c$.
		\renewcommand{\algorithmicrequire}{\textbf{Output:}}
		\REQUIRE An optimal path that covers the whole free space.
		\FOR {cohort $A_c \in A$}
		\FOR {ant $a_{l_c} \in A_c$}
		
		\STATE Choose the next cell $m_t$ according to equation (\ref{eq:ACO}).

		\STATE Update the pheromone trail according to equation (\ref{eq:localPheromone}).
		
		\STATE Repeat the above two steps until all free cells are visited.

		\ENDFOR
		\STATE Choose the best path found and deposit extra pheromone according to equation (\ref{eq:globalPheromone}).
		\ENDFOR
		\STATE Choose and return the optimal path found.
	\end{algorithmic}
\end{algorithm}

\section{Performance Validation and Analysis}
\begin{figure}[t]
    \centering
    \includegraphics[width=1.0\linewidth]{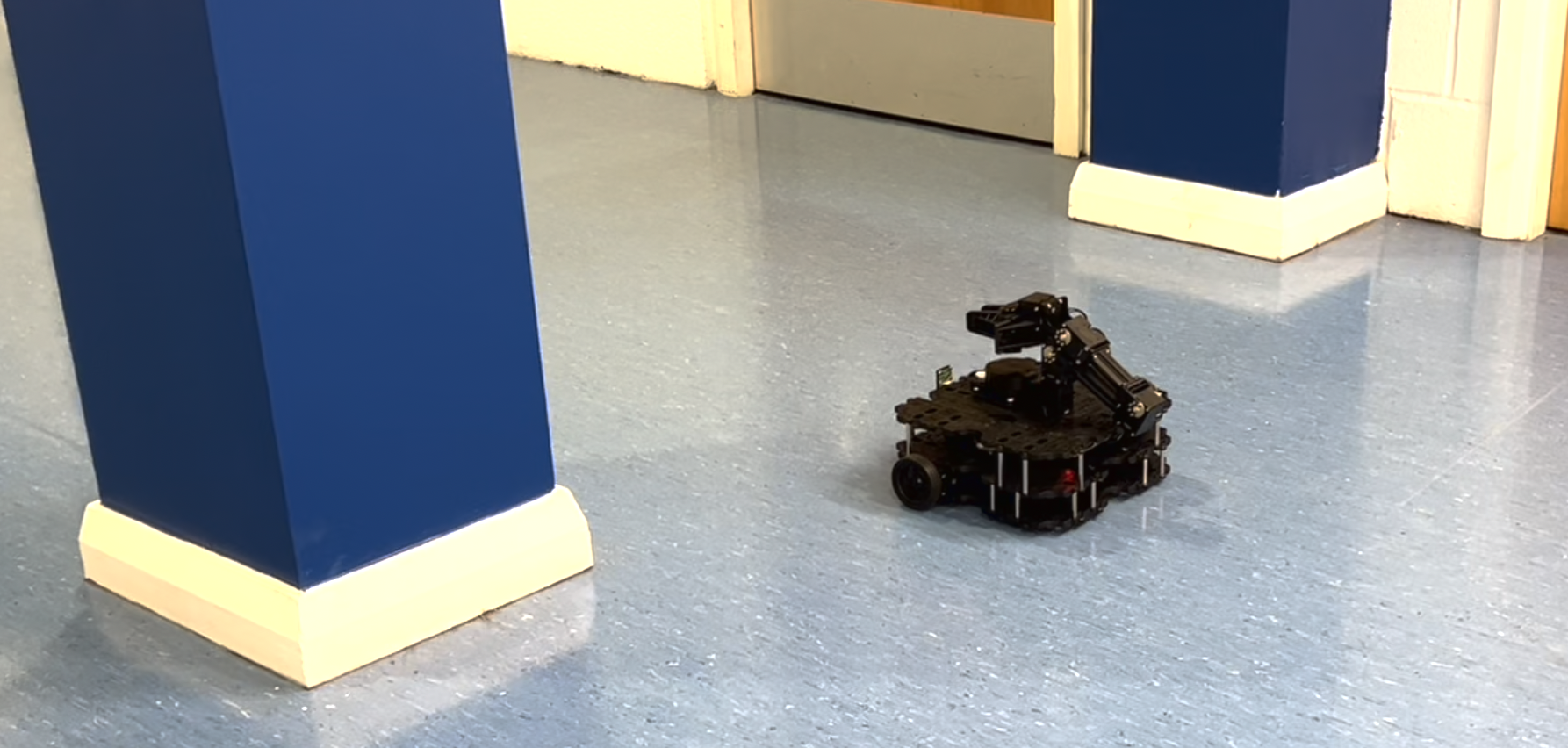}
    \caption{TurtleBot3 Waffle Pi Robot and the Office Environment}
    \label{fig:env_robot}
\end{figure}

Three different environmental maps were used for CPP to validate the performance of the proposed FaSACO in comparison with ACO, Spiral-STC, and ZigZag. The first one is the map of an office in John Dalton Building at Manchester Metropolitan University. A TurtleBot3 Waffle Pi robot with an LDS-01 laser distance sensor installed was used to build the map. The environment and robot are shown in Figure \ref{fig:env_robot}. The second one is a simulated occupancy grid map, and the third one is an open-source occupancy grid map of a basement with a size of $28\times 18.5 \text{m}^2$. For comparison, two metrics are chosen to evaluate the performance of each algorithm: 1) the number of re-covered cells, denoted as $n_r$, is the number of cells that are visited (covered) by the robot twice or more times; 2) CPU time to find the path, which is denoted as $t_o$.

To better demonstrate how velocities affect the performance of FaSACO, we deliberately designed 10 experiments that fall into three groups.
\begin{itemize}
    \item \textbf{Group 1:}  FaSACO with constant velocities
    \begin{enumerate}
        \item There are 8 experiments in this group, and ants in each experiment move with a constant velocity. To be specific, ants from the first experiment move with a velocity of one cell per step, ants from the second experiment move with a velocity of 2 cells per step, and so forth.
    \end{enumerate}
    \item \textbf{Group 2:} FaSACO with increasing velocities
    \begin{enumerate}
        \item Ants are divided into 8 cohorts, each with a velocity ranging from 1 cell per step to 8 cells per step; 
        \item The cohort with a smaller velocity carries out CPP first.
    \end{enumerate}
    \item \textbf{Group 3:} FaSACO with decreasing velocities
    \begin{enumerate}
        \item Ants are divided into 8 cohorts, each with a velocity decreasing from 8 cells per step to 1 cell per step; 
        \item The cohort with a greater velocity conducts CPP first.
    \end{enumerate}  
\end{itemize}

\begin{table*}[t]
	\centering
    \caption{Experimental Results}
    \label{tab:performance}
	\begin{tabular}{|c|c|c|c|c|c|c|}
		\hline
		\multirow{2}{*}{Velocity}  & \multicolumn{3}{c|}{FaSACO $n_r$} & \multicolumn{3}{c|}{FaSACO $t_o$ ($s$)}  \\ 
		\cline{2-7}
		                &Office       &Simulated map             &Basement     &Office        &Simulated map              &Basement                   \\ 
		\hline
		1               & 34          & 80                        & 321         &5               & 96                      &140                                   	\\ 
		\hline
		2               &\textbf{30}              & 80                        & 347      &5                  &  76                     &   126                                 \\ 
		\hline 
		3             &\textbf{30}                & 82                        & 333      &\textbf{4}                  & 65                       &  115                                   \\ 
		\hline
		4             &34                & \textbf{70}                         & 305     &\textbf{4}                  & 56                  &  108                                    \\ 
		\hline
		5             &36                & 72                         & 305     &\textbf{4}                   &  52                 &  106                                    \\ 
		\hline
		6             &\textbf{30}                & 72                         & \textbf{293}     &\textbf{4}                   &  50                 &  102                                    \\ 
		\hline
		7             &42                 & 72                         & 295    &\textbf{4}                    &  \textbf{48}                 &  101                                    \\ 
		\hline
		8             &36                  & 72                         & 307    &\textbf{4}                   &  \textbf{48}                 &  \textbf{100}                                    \\ 		
		\hline
		1 $\rightarrow$ 8       &\textbf{30}               &72                       & 307      &\textbf{4}                    &  65           &   113                                   \\ 
		\hline
		8  $\rightarrow$ 1      &\textbf{30}                 & \textbf{70}                      & 299      &\textbf{4}                    &  65            & 113                                     \\ 
		\hline		
		\hline
		\multirow{2}{*}{Others}  & \multicolumn{3}{c|}{$n_r$} & \multicolumn{3}{c|}{$t_o$ ($s$)}  \\ 
		\cline{2-7}
		
	   	            &Office     &Simulated map             &Basement      &Office       &Simulated map              &Basement                        \\ 
	   	\hline
	   	FaSACO          &\textbf{30}    & \textbf{70}                    & \textbf{299}         &4                  &  65              & 113                                \\
		\hline
		ACO             &34             & 80                    & 321        &5                   &  96              & 140                                 \\
		\hline
		Spiral-STC      &47              & 109                   & 487       & $\sim$ 0.0        & $\sim$ 0.0             &0.2                                    \\ 
		\hline
		ZigZag          &45               & 113                   & 407      & $\sim$ 0.0        & $\sim$ 0.0             & $\sim$ 0.0                                    \\
		\hline
	\end{tabular}
\end{table*}

Table \ref{tab:performance} shows the experimental results. All results are generated by 1,000 ants in total. One can see when ants move with a constant velocity (the first 8 experiments), the number of re-covered cells and the CPU time vary for all three maps. For instance, the smallest numbers of re-covered cells for the simulated map and the basement map are achieved by ants with a constant velocity of 4 cells per step and 6 cells per step, respectively. While that for the office map the best results are achieved by three constant velocities, i.e., 2, 3, and 6 cells per step. Similar results are observed for the CPU time as well, e.g., the lowest CPU time for the basement map is achieved by the constant velocity 8 cells per step, and there are more than one constant velocities that achieve the lowest CPU time for both the office map and the simulated map. While we can conclude that by changing the velocity of ants, better performance has been achieved, it is still difficult to decide which velocity one should choose. It is worth noting that the best performance in terms of the number of re-covered cells and the CPU time is achieved by different velocities for all three maps. This makes it even more difficult to determine the constant velocity. 

Instead of finding the `best' constant velocity, FaSACO suggests that we should divide the ants into cohorts of different velocities, which are demonstrated by the 9-th and 10-th experiments in Table \ref{tab:performance}. We can see that FaSACO with various ant velocities either achieves the best performance or at least a trade-off between the number of re-covered cells and the CPU time. We attribute the performance improvements to the fact that faster-moving ants communicate destination and obstacle information through pheromone trails to slower ants, thus reducing the overall time for finding the (sub-)~optimal path with a small number of re-covered cells.

\begin{figure*}[ht]
    \centering
    \subfigure[FaSACO]{
        \includegraphics[width=0.20\linewidth, height=0.12\linewidth]{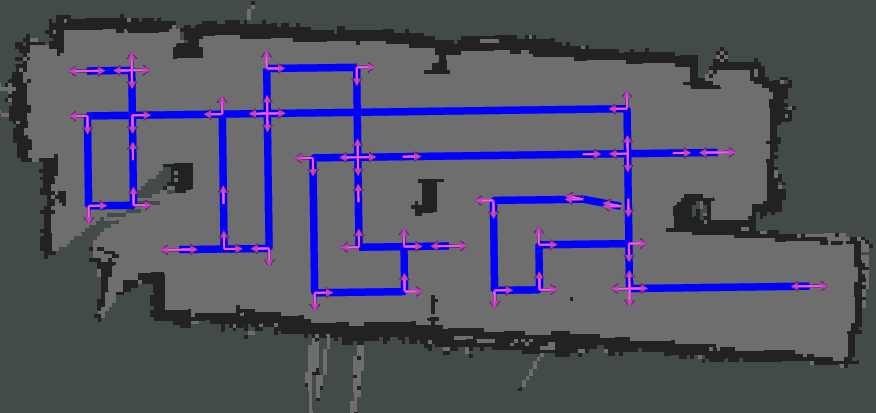}
        \label{fig:office_1}
    }
    \subfigure[ACO]{
	\includegraphics[width=0.20\linewidth, height=0.12\linewidth]{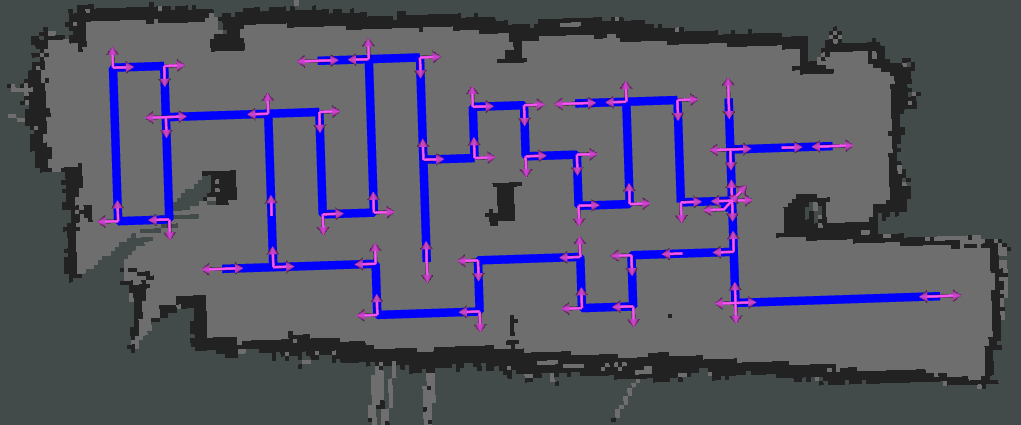}
        \label{fig:office_2}
    }
    \subfigure[Spiral-STC]{
	\includegraphics[width=0.20\linewidth, height=0.12\linewidth]{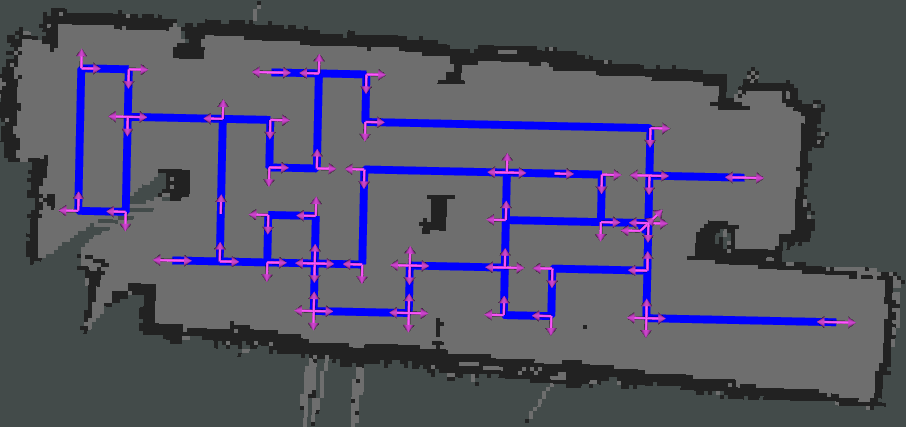}
        \label{fig:office_3}
    }    
    \subfigure[ZigZag]{
    	\includegraphics[width=0.20\linewidth, height=0.12\linewidth]{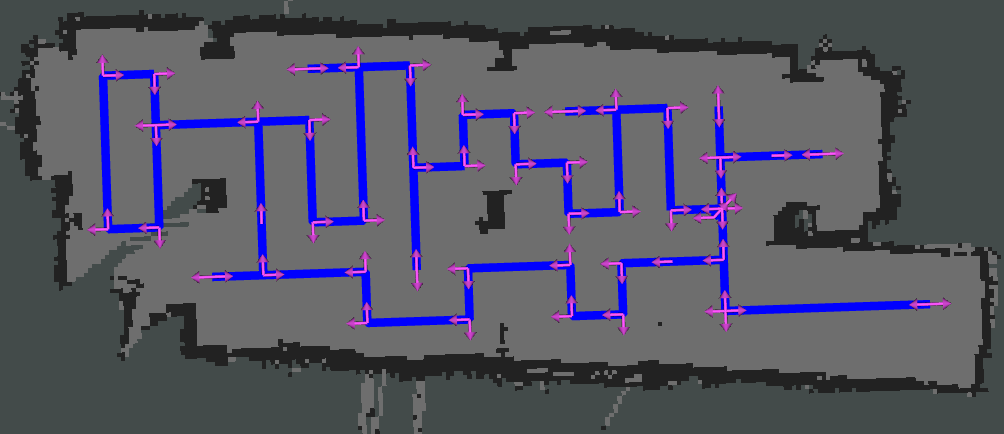}
        \label{fig:office_4}
    }
    \caption{CPP Results on the Office Map. The blue lines are path generated and the red arrows indicate robot orientations}
    \label{fig:office}
\end{figure*}

\begin{figure*}[ht]
    \centering
    \subfigure[FaSACO]{
        \includegraphics[width=0.20\linewidth, height=0.167\linewidth]{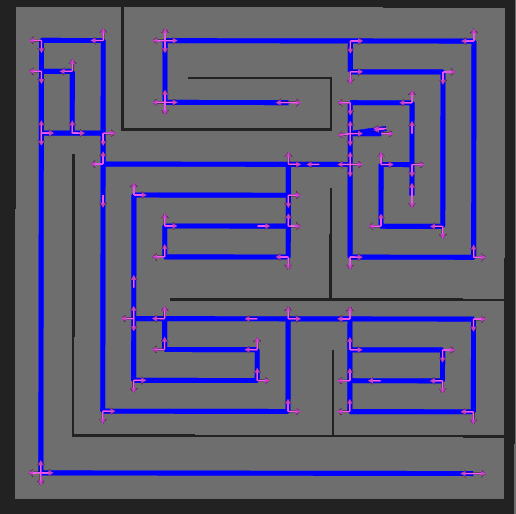}
        \label{fig:grid_1}
    }
    \subfigure[ACO]{
	\includegraphics[width=0.20\linewidth, height=0.167\linewidth]{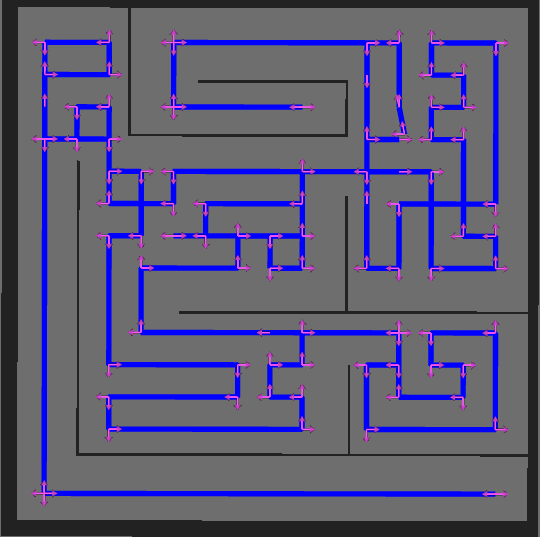}
        \label{fig:grid_2}
    }
    \subfigure[Spiral-STC]{
	\includegraphics[width=0.20\linewidth, height=0.167\linewidth]{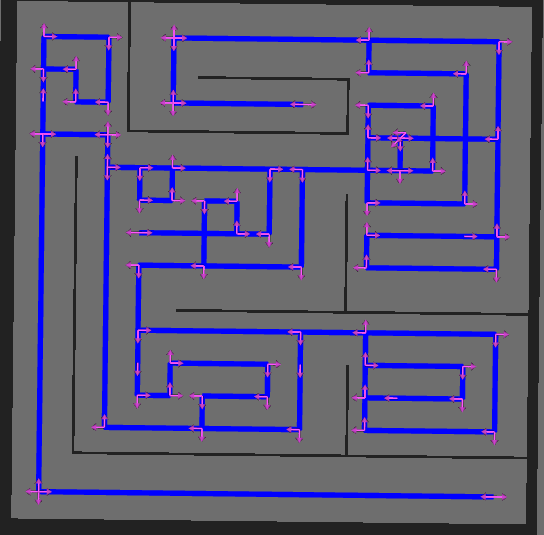}
        \label{fig:grid_3}
    }    
    \subfigure[ZigZag]{
    	\includegraphics[width=0.20\linewidth, height=0.167\linewidth]{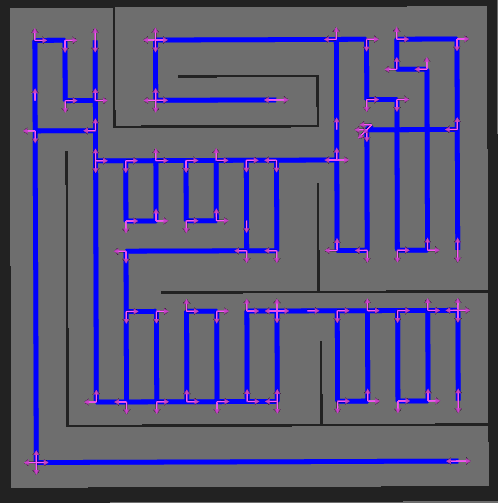}
        \label{fig:grid_4}
    }
    \caption{CPP Results on the Simulated Map. The blue lines are path generated and the red arrows indicate robot orientations}
    \label{fig:grid}
\end{figure*}

\begin{figure*}[tb]
    \centering
    \subfigure[FaSACO]{
        \includegraphics[width=0.35\linewidth, height=0.292\linewidth]{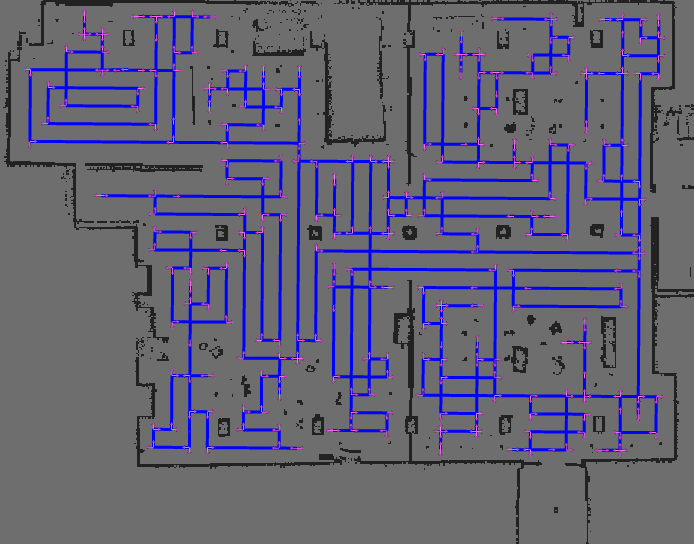}
        \label{fig:base_1}
    }
    \subfigure[ACO]{
	\includegraphics[width=0.35\linewidth, height=0.292\linewidth]{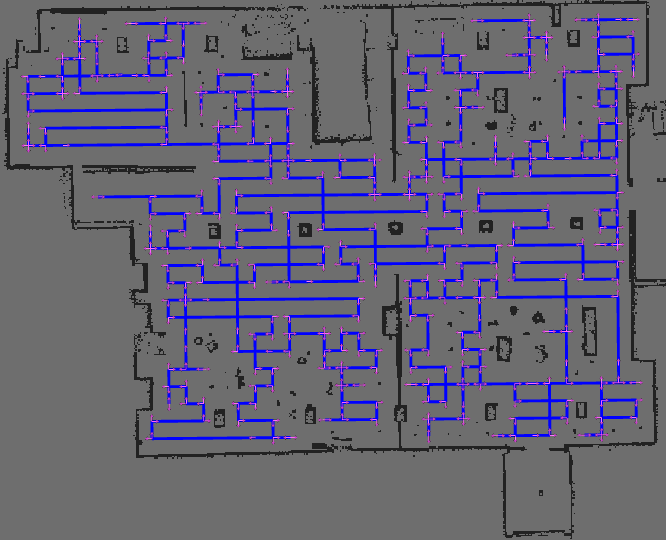}
        \label{fig:base_2}
    }
    \subfigure[Spiral-STC]{
	\includegraphics[width=0.35\linewidth, height=0.292\linewidth]{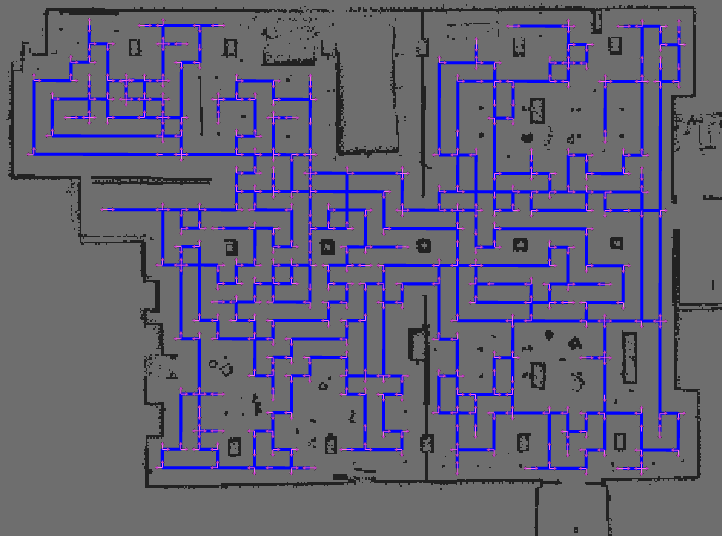}
        \label{fig:base_3}
    }    
    \subfigure[ZigZag]{
    	\includegraphics[width=0.35\linewidth, height=0.292\linewidth]{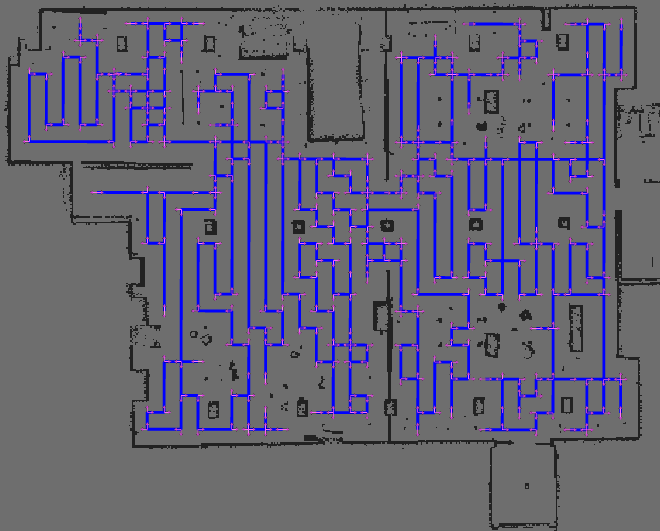}
        \label{fig:base_4}
    }
    \caption{CPP Results on the Basement Map. The blue lines are path generated and the red arrows indicate robot orientations}
    \label{fig:base}
\end{figure*}

Considering both efficacy and efficiency, the strategy in Group 3 is used in FaSACO to generate paths. The paths generated by FaSACO, ACO, Spiral-STC, and ZigZag for the three maps are given in Figures \ref{fig:office}, \ref{fig:grid}, and \ref{fig:base}. The blue lines indicate the coverage paths generated by different algorithms and the red arrows show the orientations of the robot when approaching and leaving the cell.

Overall, one can see that FaSACO outperforms ACO in all three maps, with the number of re-covered cells reduced by $6.9\% \sim 12.5\%$, and the CPU time reduced by $19.3\% \sim 32.3\%$.  When compared to Spiral-STC and ZigZag, the number of re-covered cells by FaSACO is $13.5\% \sim 38.6\%$ and $26.5\% \sim 38.1\% $ less than Spiral-STC and ZigZag, respectively. With fewer re-covered cells, FaSACO would be beneficial to applications with limited battery capacity, although FaSACO still takes more CPU time than Spiral-STC and ZigZag in the path planning stage.

\section{Conclusion}
A FaSACO-based coverage path planning algorithm is proposed. Compared to ACO, ants in FaSACO can move with different velocities. This enables ants with a higher velocity to communicate information such as destinations and obstacles to ants moving slower, making FaSACO balance between the CPU time needed for finding the path and the number of re-covered cells. Overall, FaSACO outperforms ACO both in terms of the number of re-covered cells and the CPU time consumed to find the coverage path, and outperforms Spiral-STC and ZigZag in terms of the number of re-covered cells. However, FaSACO is still less efficient than Spiral-STC and ZigZag in terms of CPU time consumed. Future work will focus on improving the efficiency of the proposed method by exploiting parallel computing techniques and searching for other methods to reduce the time complexity of FaSACO.

\bibliographystyle{IEEEtran}
\bibliography{reference}

\end{document}